\documentclass[10pt,twocolumn,letterpaper]{article}

\usepackage[pagenumbers]{cvpr} 

\usepackage{graphicx}
\usepackage{amsmath}
\usepackage{amssymb}
\usepackage{booktabs}
\usepackage{bbm}
\usepackage{arabtex}
\usepackage{utf8}
\usepackage{multirow}

\def \xa {\mathbf{x}_{adv}}
\def \xb {\mathbf{x}}
\newcommand{\argmax}{\mathop{\mathrm{arg\,max}}}
\usepackage[pagebackref,breaklinks,colorlinks]{hyperref}

\usepackage[capitalize]{cleveref}
\crefname{section}{Sec.}{Secs.}
\Crefname{section}{Section}{Sections}
\Crefname{table}{Table}{Tables}
\crefname{table}{Tab.}{Tabs.}

\def\cvprPaperID{*****} 
\def\confName{CVPR}
\def\confYear{2022}

\begin{document}
\setcode{utf8}
\title{Breaking the De-Pois Poisoning Defense}

\author{Alaa Anani\\
{\tt\small alan00001@stud.uni-saarland.de}
\and
Mohamed Ghanem\\
{\tt\small moab00005@stud.uni-saarland.de}
\and
Lotfy Abdel Khaliq\\
{\tt\small loab00001@stud.uni-saarland.de}
}
\maketitle

\begin{abstract}
   Attacks on machine learning models have been, since their conception, a very persistent and evasive issue resembling an endless cat-and-mouse game. One major variant of such attacks is poisoning attacks which can indirectly manipulate an ML model. It has been observed over the years that the majority of proposed effective defense models are only effective when an attacker is not aware of them being employed. In this paper, we show that the attack-agnostic De-Pois defense is hardly an exception to that rule. In fact, we demonstrate its vulnerability to the simplest White-Box and Black-Box attacks by an attacker that knows the structure of the De-Pois defense model. In essence, the De-Pois defense relies on a critic model that can be used to detect poisoned data before passing it to the target model. In our work, we break this poison-protection layer by replicating the critic model and then performing a composed gradient-sign attack on both the critic and target models simultaneously -- allowing us to bypass the critic firewall to poison the target model.
\end{abstract}

\section{Introduction}
\label{sec:intro}
When talking about the impact of machine learning (ML) attacks, it is important to closely consider the practical (e.g., commercial) settings into which these models are deployed. That is because many ML attacks are only applicable when certain deployment configurations are present. A good example of this is \textit{poisoning attacks} in which an attacker injects \textit{poisoned} data into an online ML model to manipulate its behavior (e.g., changing its decision boundary, etc.) \cite{jagielski2018manipulating}. Poisoning attacks essentially rely on the target model being online, i.e., it gets updated in a live fashion based on the data it receives at real-time. This online learning -- and its extended version \textit{continual learning} -- is crucial for many ML applications such as autonomous agents that interact and function in a constantly shifting and changing environment which prompts continual collection of real-time data to keep the models updated \cite{qiu2016survey}. Another primary axis to this landscape is the deployed model's utility, that is, how it is used to serve the respective clients. For an image-recognition service, the underlying model can provide its class posterior confidence vector or only the highest-likelihood class -- depending on how much information the service providers allow the model to give. This utility aspect is essential to determining the model's vulnerability to model stealing attacks, in which an attack trains a replica of the target model by using its outputs as ground-truth labels \cite{isakov2019survey}. Naturally, both dimensions -- namely online learning and model utility -- shape the preliminary context of this paper. On the one hand, the De-Pois defense is meant to prevent poisoning attacks on \textbf{online} models that operate on a static data distribution such as image classification tasks. On the other hand, deploying the De-Pois defense would introduce changes to the utility interface of the target model, namely due to the introduction of the critic model that can block incoming client-provided data. This opens the door to stealing the critic model by inferring the critic judgement either directly if the service interface returns the verdict as feedback (which is unavoidable in many applications) or indirectly by inferring the verdict via probing the model and checking whether the target effect (e.g., misclassification of certain samples) has started to manifest. Unfortunately, recent research shows that models can still be effectively stolen based on minimal-utility posteriors and with very little knowledge of the target model's network architecture thereby creating a knock-off model that performs comparably to the original model \cite{orekondy2019knockoff}. In the following sections, we shall delve deeper into the workings of the De-Pois defense model along with our corresponding attack methodology.



\section{Background}
In this section, we briefly discuss the two core preliminaries to this paper, namely, adversarial samples and the De-Pois defense model which we set out to subvert.

\subsection{Adversarial Sample Generation}

Given a pretrained ML model $h$ and a normal sample $\xb$ with class label $y$, an adversary tries to maximize the classification error of the model while keeping $\xa$ within a small $\epsilon$ at the center of the original sample $\xb$ ($\|\xa - \xb\|_p \leq \epsilon$), where $\|\cdot\|_p$ is the $L_p$-norm.

For white-box setting, adversarial samples can be generated by solving the following optimization problem:

\begin{equation}\label{eq:prob}
   \xa = \argmax_{\|\xb' - \xb\|_\infty \leq \epsilon} \ell(h(\xb'), y),
\end{equation}

where $\ell(\cdot)$ is the classification loss, and $y$ is the ground truth class. There are many different methods for generating adversarial samples in the literature . For our purposes, we employ one that is arguably the simplest: \textbf{Fast Gradient Sign Method (FGSM)}. This method works by perturbing normal examples $\xb$ for one step by the amount of $\epsilon$ along the input gradient direction to maximize the loss \cite{goodfellow2014explaining}. More formally,
\begin{equation}
    \xa = \xb + \epsilon \cdot \text{sign}(\nabla_{\xb} \ell(h(\xb), y)).
\end{equation}



\subsection{The De-Pois Model}
Most of the defenses against posioining attacks are specific to the model being attacked since poisoned data generation is dependent on the model's loss function as well as the defense mechansims the model owner is using. In De-Pois \cite{chen2021depois}, Jian Chen et al. propose a generic and attack-agnostic defense approach in which a mimic model is trained to imitate the behaviour of a target model trained on \emph{trusted} clean samples. Subsequently, a critic model is trained to filter-out poisoned samples by comparing the prediction differences between the mimic model and the target model. The De-Pois architecture can be summarized in 3 steps:
\begin{itemize}
    \item \textbf{Synthetic Data Generation}: In this step, a synthetic dataset is created using cGAN \cite{mirza2014conditional} with the same distribution as a trusted clean dataset $S_{c}$. The generated samples are conditioned on the class label and an authenticator \cite{tran:2017bayesian} is employed to supervise the data augmentation process. At the end, $S_{aug}$ is created by augmenting the generated samples with the clean samples.
    \item \textbf{Mimic Model Construction}  De-Pois next builds the mimic model using WGAN-GP \cite{gulrajani2017improved} to learn the distribution of predictions of the augmented training data. When the training process is completed, WGAN-GP's discriminator is regarded as the critic model.
    \item \textbf{Poisoned Data Recognition (Inference)}: Prior to inference, a decision boundary is derived from the trusted dataset which is used to distinguish between clean and poisoned data. When an image is passed to the critic model, this decision boundary is employed to detect poisoning. If the critic model’s output score is lower than than the detection boundary, the sample is regarded as poisoned.

\end{itemize}


\label{sec:formatting}

\section{Methodology}

\begin{figure*}
    \centering
    \includegraphics[height=0.35\textwidth, width=0.9\textwidth]{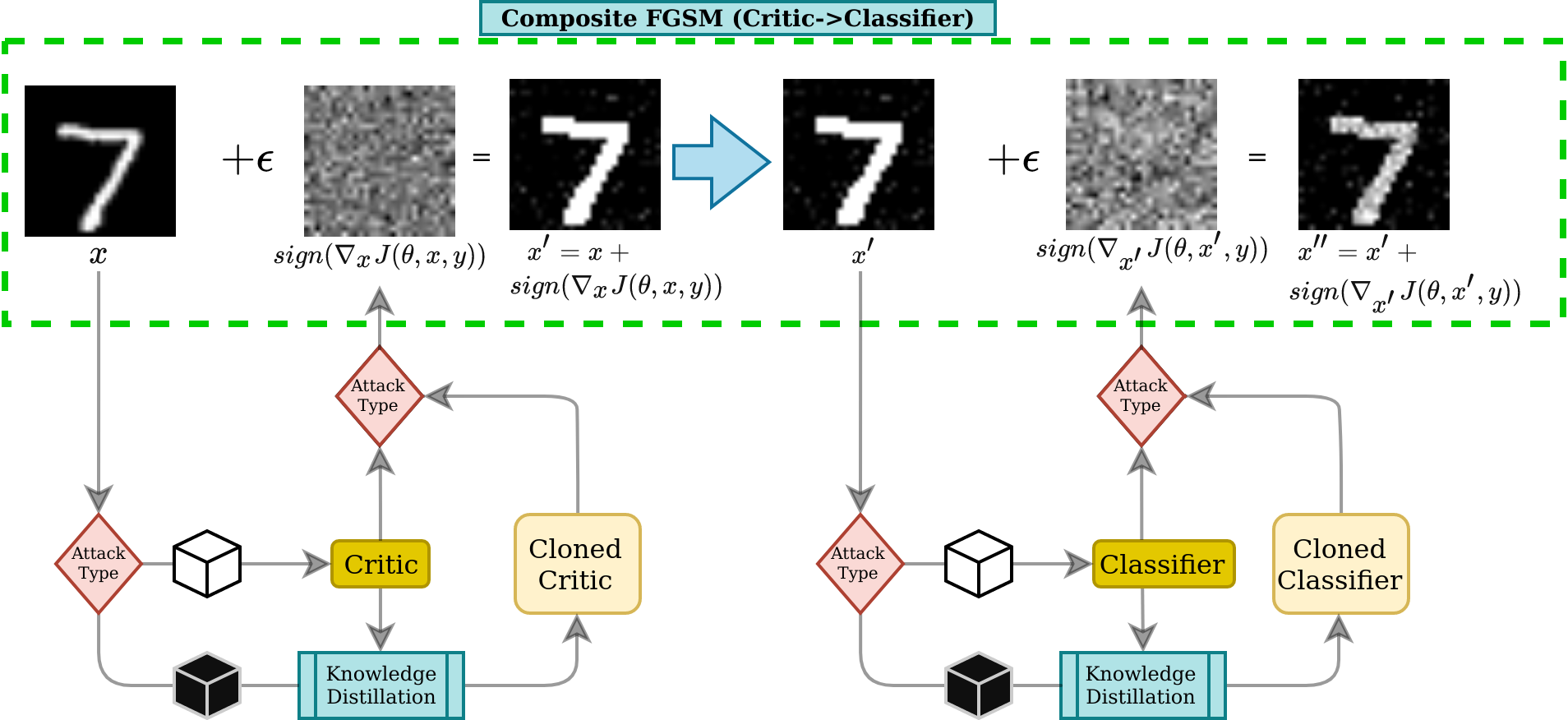}
    \caption{The Overall Attack on De-Pois Flow Including the Composite FGSM Module in Both the White-Box and the Black-Box Attack}
    \label{fig:diagram}
\vspace{0.1cm}
\end{figure*}

We attack the De-Pois model in two different access modes: white-box and black-box. In the first mode, we perform a white-box attack using FGSM directly on both the critic and the classifier. In the second mode, we create shadow models via knowledge distillation for both the critic and the classifier, and attack them mode using FGSM to generate adversarial samples against the original models. The reason why we conducted the white-box attack first is to have a baseline efficacy to compare with the shadow models in the black-box mode. Figure \ref{fig:diagram} shows the overall attack structure. 
\subsection{White-Box FGSM Attack}
We use FGSM in a compositional manner. Given a clean sample $x_i$, we get a critic-poisoned sample by adding the critic's signed gradients to $x_i$ (i.e., generating $\textit{FGSM}_{critic}(x_i)$). Afterwards, we add the signed gradients of the classifier w.r.t $\textit{FGSM}_{critic}(x_i)$, effectively generating the doubly-poisoned sample: 
\[{x_p}_i = \textit{FGSM}_{classifier}(\textit{FGSM}_{critic}(x_i))\]

We also experiment with the reverse order of the composition (i.e., $\textit{FGSM}_{critic}(\textit{FGSM}_{classifier}(x_i))$) and the usage of singly-poisoned samples $\textit{FGSM}_{critic}(x_i)$ and $\textit{FGSM}_{classifier}(x_i)$ to find which attack order is more effective.

\subsection{Black-Box Attack with Knowledge Distillation}
To mount a black-box attack (i.e., accessing only the target model's posteriors), we perform knowledge distillation on both the critic and the classifier. We take the trained critic and classifier models as teacher models and transfer their knowledge to two student models by minimizing a loss function in which the target is the posteriors of the teacher model. We also employ a \textit{softmax temperature} introduced by Hinton et al. \cite{hinton2015distilling} to mitigate the high predicted class probabilities in the teacher posteriors. The rationale behind this is to smoothen the target model posteriors to make it more informative to the student model. We minimize the following objective:
\begin{equation}
    L = \alpha * L_{s} + \mathbb{KL}(p_{t}||p_{s})
\end{equation}
where $L_{s}$ is the student loss, $\mathbb{KL}(\cdot)$ is the Kullback–Leibler divergence, $p_{t}$ is the teacher's posteriors, $p_{s}$ is the student's posteriors, and $\alpha$ is a hyperparameter, which we set to 0.5. Hence, we obtain shadow models that mimic the behavior of the target critic and classifier models. Using these shadow models, we conduct our attack, which should typically allow us to obtain an estimate of the signed gradients on the original models thereby creating an adversarial dataset to attack the original models in succession.

\section{Results}
\subsection{Experiments Setup}
The goal behind our attacks is to hinder the performance of the De-Pois model by supplying it with adversarial data, which we define by two metrics: the critic's accuracy (ca) and the overall De-Pois accuracy (da). For a poisoned sample ${x_p}_i = (x_i, y_i)$ we re-define its ground truth $t_i=\{-1, y_i\}$ where $-1$ denotes that ${x_p}_i$ is a poisoned sample and $y_i$ is the correct class for the image $x_i$. In other words, an image is deemed correctly classified by the combined De-Pois model if it was either dismissed by the critic as poisoned or allowed by the critic and then correctly classified by the classifier since that counts as an overall positive. Given this ground truth, we define the critic's accuracy ($ca$) as follows:
\[ca = \frac{\sum^{|P|}_{i=0} \mathbbm{1}(\hat{t}_{i} = -1)}{|P|} \]
where $\mathbbm{1}$ is the index function that returns 1 on true, and 0 otherwise, and $\hat{t}_i$ is the De-Pois prediction of the sample ${x_p}_i$. Note that De-Pois returns $\hat{t}_i=-1$ if and only if the sample ${x_p}_i$ is detected as poisoned (hence, rejected) by the critic. To measure the overall performance of the De-Pois pipeline against our attack, we define the overall metric, De-Pois Accuracy (da) to include both correctly rejected samples along with correctly classified samples as follows:
\[da = \frac{\sum^{|P|}_{i=0} \mathbbm{1}(\hat{t}_{i} \in t_i)}{|P|} \].

In total, we experiment with 4 different attack modes/orders. We take the \textit{classifier-only} mode as our baseline since this is the de facto attack mode performed by an attacker who is unaware of the critic model. For simplicity, we conduct our experiments in the scope of the MNIST hand-written digits dataset \cite{deng2012mnist}, but the analysis is further applicable to other datasets.


\subsection{White-Box Attack}

\begin{figure}[!hbtp]
    \centering
    \includegraphics[width=0.4\textwidth]{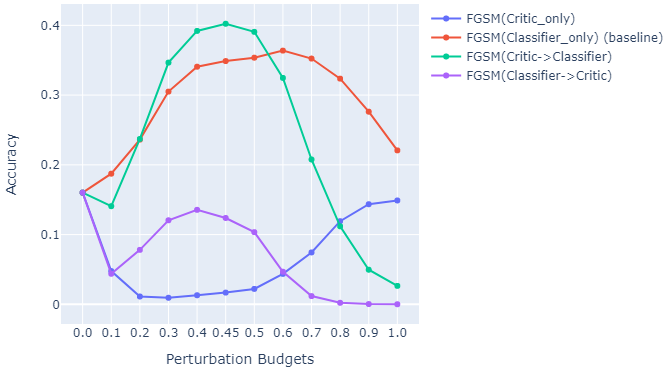}
    \caption{Critic Accuracy (ca) vs. Perturbation Budget (White-Box) Across 4 Attack Modes}
    \label{fig:cawb}
\end{figure}
In Figure \ref{fig:cawb}, there is an overall pattern in the accuracy behavior in all attack modes except critic-only. In critic-only, the accuracy initially decreases dramatically until epsilon 0.3, then it starts to increase by increasing epsilon. The reason why this happens is because FGSM is applied directly using the critic's gradients, so it effectively fools the critic until a certain threshold (0.3), after which the gradients start to make the image look poisoned/suspicious to the critic, hence the critic detects more poisoned images. Regarding the result of the attack modes, the gradients in this case are either not from the critic (Classifier-only) or from a combination of both the critic and the classifier (Classifier-Critic and Critic-Classifier). The accuracy across epsilons follows a trend of a parabola-like behavior by increasing at first and then decreasing. The first increasing half of the graph is explained by the fact that the gradients are not specifically designed enough to trick the critic, hence it can detect many images as poisoned. In the second half, specifically after epsilon is larger than 0.4, a domain-shift occurs from the kind of images the critic is trained to detect to ones that look obviously poisoned due to the high perturbation noise, but this is not detected by the critic.

\begin{figure}[!hbtp]
    \centering
    \includegraphics[width=0.4\textwidth]{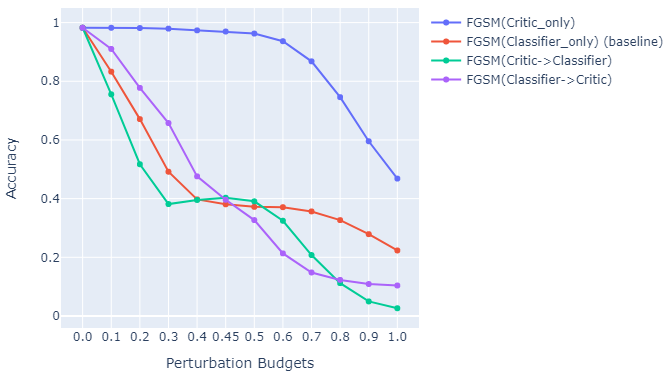}
    \caption{De-Pois Accuracy (da) vs. Perturbation Budget Across 4 Attack Modes (White-Box)}
    \label{fig:dawb}
\end{figure}
In Figure \ref{fig:dawb}, the most successful attack mode is Critic-Classifier, scoring an accuracy of $<40\%$ at epsilon $0.3$. The rest of the attack modes decrease the accuracy, but not as fast or as low as the best mode. From epsilon 0.3 to 0.7, we can witness the effect of the previously described parabola in the accuracy of the critic as the attack does not improve by increasing epsilon. However, afterwards, the critic accuracy starts to drop (alongside the classifier's), which hinders the overall pipeline accuracy of De-Pois.

\subsection{Black-Box Attack}

\begin{figure}[!htbp]
    \centering
    \includegraphics[width=0.4\textwidth]{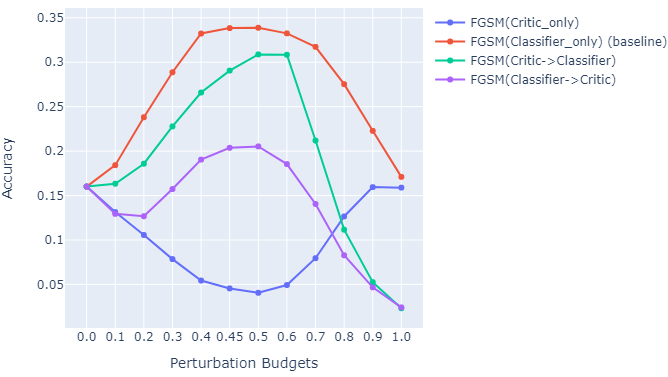}
    \caption{Critic Accuracy (ca) vs. Perturbation Budget Across 4 Attack Modes (Black-Box)}
    \label{fig:cabb}
\end{figure}

In Figure \ref{fig:cabb}, we detect in the Black-Box attack a similar trend for the critic accuracy across the 4 attack modes in the White-Box attack. This evidently shows that the shadow models effectively mirror the original models as desired.

\begin{figure}[!htbp]
    \centering
    \includegraphics[width=0.4\textwidth]{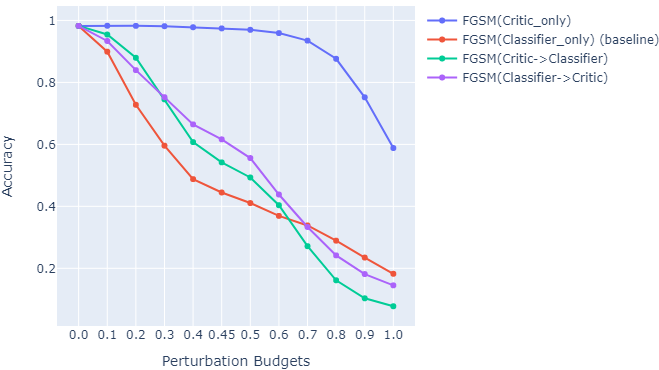}
    \caption{De-Pois Accuracy (da) vs. Perturbation Budget (Black-Box)}
    \label{fig:dabb}
\end{figure}

In Figure \ref{fig:dabb}, the De-Pois accuracy follows a similar trend in the Black-Box attack to the White-Box attack. However, the best attack mode is (Classifier-only), as the De-Pois accuracy increases faster in all epislons until ~0.7, then the Critic-Classifier mode decreases the accuracy further. 

\subsection{Overall Comparison}
\begin{figure}[htbp!]
    \centering
    \includegraphics[width=0.4\textwidth]{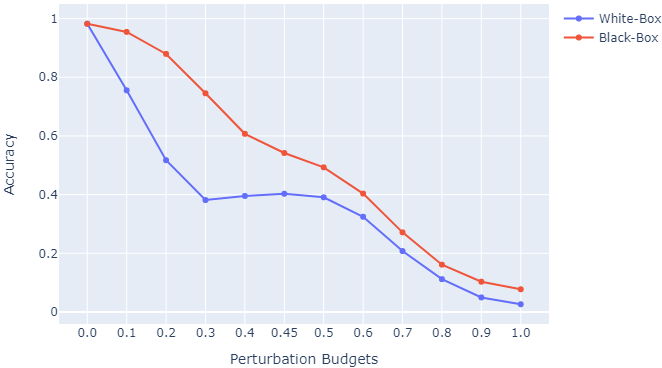}
    \caption{Overall Comparison between the De-Pois Accuracy across the White-Box and Black-Box attacks}
    \label{fig:overall}
\end{figure}

We observe the attack mode Critic-Classifier to be the most effective based on the previous case-analysis of the results for both the White-Box and the Black-Box attacks. Hence, we use the Critic-Classifier mode in the overall comparison between the White-Box and the Black-Box access modes in terms of overall De-Pois accuracy in Figure \ref{fig:overall}. 

\begin{table}[!htbp]
  \centering
  \begin{tabular}{l l c c}
    \toprule
    \multicolumn{1}{p{1cm}}{\centering \textbf{Access} \\  \textbf{Mode}} & \textbf{Target} & \multicolumn{1}{p{1cm}}{\centering \textbf{Critic} \\  Acc.} & \multicolumn{1}{p{1.5cm}}{\centering \textbf{De-Pois} \\  Acc.}  \\
    \midrule
    \multirow{2}{*}{White-box} & Classifier-Only & 0.3524 & 0.3564 \\
    & Composed & \textbf{0.2077} & \textbf{0.2077} \\ \hline
    \multirow{2}{*}{Black-box} & Classifier-Only & 0.3173 & 0.3383 \\ 
    &Composed & \textbf{0.2119} & \textbf{0.2714} \\
    \bottomrule
  \end{tabular}
  \caption{Overall Results}
  \label{tab:overall}
\end{table}

We choose the smallest effective perturbation budget for the attack to be 0.7 where the De-Pois accuracy in the black-box attack is the closest to the white-box because larger budgets could disfigure the original images. All in all, the black-box attack effectively decreases the De-Pois accuracy from 100\% to $\sim40\%$ ($\sim 60\%$ decrease) compared to the $\sim70\%$ decrease of the White-Box attack. Table \ref{tab:overall} shows the highlight results.

\section{Conclusion}
In summary, in this paper, we have demonstrated that the De-Pois defense model can be broken with a combination of techniques as simple as knowledge distillation and FGSM. We have further explained the reason behind this vulnerability being that the critic model is prone to stealing attacks which compromise its protection layer since it is vulnerable -- like almost all ML models -- to evasion attacks. As a result, the target model is once again open to poisoning.

{\small
\bibliographystyle{ieee_fullname}
\bibliography{PaperForReview}
}

\end{document}